\documentclass{article}
\usepackage{iclr2016_conference}
\usepackage{times}
\usepackage{hyperref}
\usepackage{url}
\usepackage{natbib}

\usepackage{algorithm}
\usepackage{algorithmic}
\usepackage{amsmath,amssymb,amsthm}
\usepackage{psfrag}
\usepackage{color}
\usepackage{pdfpages}
\usepackage{wrapfig}
\iclrfinalcopy

\newcommand{\A}{\ensuremath{\mathbf{A}}}
\newcommand{\B}{\ensuremath{\mathbf{B}}}
\newcommand{\C}{\ensuremath{\mathbf{C}}}

\newcommand{\I}{\ensuremath{\mathbf{I}}}

\newcommand{\K}{\ensuremath{\mathbf{K}}}

\newcommand{\RR}{\ensuremath{\mathbf{R}}}
\renewcommand{\SS}{\ensuremath{\mathbf{S}}}
\newcommand{\T}{\ensuremath{\mathbf{T}}}
\newcommand{\U}{\ensuremath{\mathbf{U}}}
\newcommand{\V}{\ensuremath{\mathbf{V}}}

\newcommand{\X}{\ensuremath{\mathbf{X}}}
\newcommand{\Y}{\ensuremath{\mathbf{Y}}}

\newcommand{\f}{\ensuremath{\mathbf{f}}}
\newcommand{\g}{\ensuremath{\mathbf{g}}}

\newcommand{\uu}{\ensuremath{\mathbf{u}}}
\newcommand{\vv}{\ensuremath{\mathbf{v}}}
\newcommand{\w}{\ensuremath{\mathbf{w}}}
\newcommand{\x}{\ensuremath{\mathbf{x}}}
\newcommand{\y}{\ensuremath{\mathbf{y}}}

\newcommand{\0}{\ensuremath{\mathbf{0}}}

\newcommand{\balpha}{\ensuremath{\boldsymbol{\alpha}}}
\newcommand{\bbeta}{\ensuremath{\boldsymbol{\beta}}}

\newcommand{\bLambda}{\ensuremath{\boldsymbol{\Lambda}}}

\newcommand{\bSigma}{\ensuremath{\boldsymbol{\Sigma}}}

\newcommand{\bbR}{\ensuremath{\mathbb{R}}}

\newcommand{\calN}{\ensuremath{\mathcal{N}}}
\newcommand{\calO}{\ensuremath{\mathcal{O}}}

\newcommand{\abs}[1]{\left\lvert#1\right\rvert}
\newcommand{\norm}[1]{\left\lVert#1\right\rVert}

{%
\begin{list}{#1}{
\vspace{-\topsep}
\vspace{-\partopsep}
\setlength{\itemindent}{0cm}
\setlength{\rightmargin}{0cm}
\setlength{\listparindent}{0cm}
\settowidth{\labelwidth}{#1}
\setlength{\leftmargin}{\labelwidth}
\addtolength{\leftmargin}{\labelsep}
\setlength{\itemsep}{0cm}
}%
}%
{%
\end{list}
\vspace{-\topsep}
\vspace{-\partopsep}
}

{\begin{enumerate}%
}%
{\end{enumerate}}

\newcommand{\traceop}{\operatorname{tr}}
\newcommand{\trace}[1]{\ensuremath{\traceop\left(#1\right)}}

\theoremstyle{plain}

\newtheorem*{lemma*}{Lemma}

\newtheorem*{prop*}{Proposition}

\theoremstyle{definition}

\newtheorem*{defn*}{Definition}

\newtheorem*{exmp*}{Example}

\newtheorem*{conj*}{Conjecture}

\theoremstyle{remark}

\newtheorem*{rmk*}{Remark}

\newcommand{\karen}[1]{}
\newcommand{\weiran}[1]{}

\title{Large-Scale Approximate Kernel Canonical Correlation Analysis}

\author{Weiran Wang \hspace{1em} \&  \hspace{1em} Karen Livescu \\
Toyota Technological Institute at Chicago \\
6045 S. Kenwood Ave., Chicago, IL 60637 \\
Email: \{weiranwang,klivescu\}@ttic.edu}

\begin{document}

\maketitle

\begin{abstract}

Kernel canonical correlation analysis (KCCA) is a 
nonlinear multi-view representation learning technique with broad applicability in statistics and machine learning. Although there is a closed-form solution for the KCCA objective, it involves solving an $N\times N$ eigenvalue system where $N$ is the training set size, making its computational requirements in both memory and time prohibitive for large-scale problems.
Various approximation techniques have been developed for KCCA.  A commonly used approach is to first transform the original inputs to an $M$-dimensional random feature space 
so that inner products in the feature space approximate kernel evaluations, and then apply linear CCA to the transformed inputs. In many applications, however, the dimensionality $M$ of the random feature space may need to be very large in order to obtain a sufficiently good approximation; it then becomes challenging to perform the linear CCA step on the resulting very high-dimensional data matrices.
We show how to use a stochastic optimization algorithm, recently proposed for linear CCA and its neural-network extension, to further alleviate the computation requirements of approximate KCCA. This approach allows us to run approximate KCCA on a speech dataset with $1.4$ million training samples and a random feature space of dimensionality $M=100000$ on a typical workstation.
\end{abstract}

\section{Introduction}

Canonical correlation analysis (CCA, \citealp{Hotell36a}) and its extensions are ubiquitous techniques in scientific research areas for revealing the common sources of variability in multiple views of the same phenomenon, including meteorology~\citep{Anders03a}, chemometrics~\citep{Montan_95a}, genomics~\citep{Witten_09a}, computer vision~\citep{Kim_07c,SocherLi10a}, speech recognition~\citep{Rudzic10a,AroraLivesc13a,Wang_15a}, and natural language processing~\citep{Vinokour_03a,Haghig_08a,Dhillon_11b,Hodosh_13a,FaruquiDyer14a,Lu_15a}.
CCA seeks linear projections of two random vectors (views), such that the resulting low-dimensional vectors are maximally correlated. Given a dataset of $N$ pairs of observations ${(\x_1,\y_1),\dots,(\x_N,\y_N)}$ of the random variables, where $\x_i\in \bbR^{d_x}$ and $\y_i\in \bbR^{d_y}$ for $i=1,\dots,N$, the objective of CCA for $L$-dimensional projections can be written as\footnote{In this paper, we assume that the inputs are centered at the origin for notational simplicity; if they are not, we can center them as a pre-processing operation.} (see, e.g., \citealp{Borga01a})
 \begin{gather} \label{e:cca}
\max_{\U\in \bbR^{d_x\times L},\V\in \bbR^{d_y\times L}} \quad  \trace{\U^\top \bSigma_{xy} \V} \\
\text{s.t.} \quad \U^\top \bSigma_{xx} \U = \V^\top \bSigma_{yy} \V = \I,  \qquad
\uu_i^\top \bSigma_{xy} \vv_j=0, \; \text{for}\; i\neq j, \nonumber
\end{gather}
where $(\U,\V)$ are the projection matrices for each view, $\bSigma_{xy}=\frac{1}{N} \sum_{i=1}^N \x_i \y_i^\top$, $\bSigma_{xx}=\frac{1}{N} \sum_{i=1}^N \x_i \x_i^\top  + r_x \I$, $\bSigma_{yy}=\frac{1}{N} \sum_{i=1}^N \y_i \y_i^\top + r_y \I$ are the cross- and auto-covariance matrices, and $(r_x,r_y)\ge 0$ are regularization parameters~\citep{Vinod76a,BieMoor03a}. There exists a closed-form solution to \eqref{e:cca} as follows. Let the rank-$L$ singular value decomposition (SVD) of the whitened covariance matrix $\T=\bSigma_{xx}^{-\frac{1}{2}} \bSigma_{xy} \bSigma_{yy}^{-\frac{1}{2}} \in \bbR^{d_x\times d_y}$ be $\tilde{\U}\Lambda \tilde{\V}^\top$, where $\Lambda$ contains the top $L$ singular values $\sigma_1\ge \dots \ge \sigma_L$ on its diagonal, and $(\tilde{\U}, \tilde{\V})$ are the corresponding singular vectors. Then the optimal projection matrices $(\U,\V)$ in \eqref{e:cca} are $(\bSigma_{xx}^{-\frac{1}{2}} \tilde{\U}, \bSigma_{yy}^{-\frac{1}{2}} \tilde{\V})$, and the optimal objective value, referred to as the canonical correlation, is $\sum_{l=1}^L \sigma_l$.\footnote{Alternatively, one could also solve some equivalent $M\times M$ eigenvalue system instead of the SVD of $\T$, at a similar cost.} The theoretical properties of CCA~\citep{KakadeFoster07a,Chaudh_09a,Foster_09a} and its connection to other methods~\citep{Borga01a,BachJordan05a,Chechik_05a} have also been studied.

One limitation of CCA is its restriction to linear mappings, which are often insufficient to reveal the highly nonlinear relationships in many real-world applications. To overcome this issue, kernel CCA (KCCA) was proposed indepedently by several researchers~\citep{LaiFyfe00a,Akaho01a,Melzer_01a} and has become a common technique in statistics and machine learning~\citep{BachJordan02a,Hardoon_04a}. KCCA extends CCA by mapping the original inputs in both views into reproducing kernel Hilbert spaces (RKHS) and solving linear CCA in the RKHS.  By the representer theorem of RKHS~\citep{SchoelSmola01a}, one can conveniently work with the kernel functions instead of the high-dimensional (possibly infinite-dimensional) RKHS, and the projection mapping is a linear combination of kernel functions evaluated at the training samples. KCCA has been successfully used for cross-modality retrieval~\citep{Hardoon_04a,LiShawe-Taylor05a,SocherLi10a,Hodosh_13a}, acoustic feature learning~\citep{AroraLivesc13a}, computational biology~\citep{Yamanis_04a,Hardoon_07a,Blasch_11a}, and statistical independence measurement~\citep{BachJordan02a,Fukumiz_07a,Lopez_13a}.

KCCA also has a closed-form solution, via an $N \times N$ eigenvalue system (see Sec.~\ref{s:kcca}).  However, this solution does not scale up to datasets of more than a few thousand training samples, due to the time complexity of solving the eigenvalue system ($\calO(N^3)$ for a naive solution) and the memory cost of storing the kernel matrices. As a result, various approximation techniques have been developed, most of which are based on low-rank approximations of the kernel matrices. With rank-$M$ approximations of the kernel matrices, the cost of solving approximate KCCA reduces to $\calO(M^2 N)$ (see, e.g.,~\citealp{BachJordan02a,Lopez_14b}). Thus if $M \ll N$, the approximation leads to significant computational savings.  Typically, ranks of a few hundred to a few thousand are used for the low-rank kernel approximations~\citep{Yang_12e,Le_13a,Lopez_14b}. In more challenging real-world applications, however, it is observed that the rank $M$ needed for an approximate kernel method to work well can be quite large, on the order of tens or hundreds of thousands (see \citealp{Huang_14a,Lu_15b} for classification tasks, and \citealp{Wang_15a} for KCCA). In such scenarios, it then becomes challenging to solve even approximate KCCA.

In this paper, we focus on the computational challenges of scaling up approximate kernel CCA using low-rank kernel approximations when both the training set size $N$ and the approximation rank $M$ are large. The particular variant of approximate KCCA we use, called randomized CCA~\citep{Lopez_14b}, transforms the original inputs to an $M$-dimensional feature space using random features~\citep{RahimiRecht08a,RahimiRecht09a} so that inner products in the new feature space approximate the kernel function.  This approach thus turns the original KCCA problem into a very high-dimensional linear CCA problem of the form~\eqref{e:cca}. We then make use of a stochastic optimization algorithm, recently proposed for linear CCA and its deep neural network extension deep CCA~\citep{Ma_15b,Wang_15c}, to reduce the memory requirement for solving the resulting linear CCA problem.  This algorithm updates parameters iteratively based on small minibatches of training samples. This approach allows us to run approximate KCCA on an $8-$million sample dataset of MNIST digits, and on a speech dataset with $1.4$ million training samples and rank (dimensionality of random feature space) $M=100000$ on a normal workstation. Using this approach we achieve encouraging results for multi-view learning of acoustic transformations for speech recognition.\footnote{Our MATLAB implementation is available at \texttt{http://ttic.uchicago.edu/\~{}wwang5/knoi.html}} \karen{mention the name NOI explicitly?} \weiran{In that case we need to mention AppGrad as well.} \karen{true.  That would be fine.  I mildly prefer mentioning both but I am also OK with it as is.}

In the following sections we review approximate KCCA and random features (Sec.~\ref{s:kcca}), present the stochastic optimization algorithm (Sec.~\ref{s:sto}), discuss related work (Sec.~\ref{s:related}), and demonstrate our algorithm on two tasks (Sec.~\ref{s:expt}).

\section{Approximate KCCA}
\label{s:kcca}

\subsection{KCCA solution}

In KCCA, we transform the inputs $\{\x_i\}_{i=1}^N$ of view 1 and $\{\y_i\}_{i=1}^N$ of view 2 using feature mappings $\phi_x$ and $\phi_y$ associated with some positive semi-definite kernels $k_x$ and $k_y$ respectively, and then solve the linear CCA problem~\eqref{e:cca} for the feature-mapped inputs~\citep{LaiFyfe00a,Akaho01a,Melzer_01a,BachJordan02a,Hardoon_04a}. The key property of such kernels is that $k_x(\x,\x')=<\phi_x(\x),\phi_x(\x')>$ (similarly for view 2)~\citealp{SchoelSmola01a}. Even though the feature-mapped inputs live in possibly infinite-dimensional RKHS, replacing the original inputs $(\x_i, \y_i)$ with $(\phi_x(\x_i), \phi_y(\y_i))$ in \eqref{e:cca}, and using the KKT theorem~\citep{NocedalWright06a}, one can show that the solution has the form $\U=\sum_{i=1}^N {\phi_x(\x_i) \balpha_i^\top}$ and $\V=\sum_{i=1}^N {\phi_y(\y_i) \bbeta_i^\top}$ where $\balpha_i, \bbeta_i \in \bbR^L$, $i=1,\dots,N$, as a result of the representer theorem~\citep{SchoelSmola01a}.   The final KCCA projections can therefore be written as ${\f}(\x)=\sum_{i=1}^N {\balpha_i k_x(\x,\x_i)}\in \bbR^L$ and ${\g}(\y)=\sum_{i=1}^N {\bbeta_i k_y(\y,\y_i)}\in \bbR^L$ for view 1 and view 2 respectively.

Denote by $\K_x$ the $N \times N$ kernel matrix for view 1, i.e., $(\K_x)_{ij}=k_x (\x_i, \x_j)$, and similarly denote by $\K_y$ the kernel matrix for view 2. Then \eqref{e:cca} can be written as a problem in the coefficient matrices $\A=[\balpha_1,\dots,\balpha_N]^\top \in \bbR^{N\times L}$ and $\B=[\bbeta_1,\dots,\bbeta_N]^\top \in \bbR^{N\times L}$.  One can show that the optimal coefficients $\A$ correspond to the top $L$ eigenvectors of the $N\times N$ matrix $(\K_x + N r_x\I)^{-1} \K_y (\K_y + N r_y\I)^{-1} \K_x$, and a similar result holds for $\B$ (see, e.g., \citealp{Hardoon_04a}). This involves solving an eiaylorgenvalue problem of size $N\times N$, which is expensive both in memory (storing the kernel matrices) and in time (solving the $N\times N$ eigenvalue systems naively costs $\calO(N^3)$).

Various kernel approximation techniques have been proposed to scale up KCCA, including Cholesky decomposition~\citep{BachJordan02a}, partial Gram-Schmidt~\citep{Hardoon_04a}, and incremental SVD~\citep{AroraLivesc12a}. Another widely used approximation technique for kernel matrices is the Nystr\"om method~\citep{WilliamSeeger01a}. In the Nystr\"om method, we select $M$ (random or otherwise) training samples $\tilde{\x}_1,\dots,\tilde{\x}_M$ and construct the $M\times M$ kernel matrix $\tilde{\K}_x$ based on these samples, i.e.~$(\tilde{\K}_x)_{ij}=k_x(\tilde{\x}_i,\tilde{\x}_j)$.  We compute the eigenvalue decomposition $\tilde{\K}_x=\tilde{\RR} \tilde{\bLambda} \tilde{\RR}^\top$, and then the $N\times N$ kernel matrix for the entire training set can be approximated as $\K_x \approx \C \tilde{\K}_x^{-1} \C^\top$ where $\C$ contains the columns of $\K_x$ corresponding to the selected subset, i.e., $\C_{ij}=k_x(\x_i,\tilde{\x}_j)$. This means $\K_x \approx (\C \tilde{\RR} \tilde{\bLambda}^{-\frac{1}{2}}) (\C \tilde{\RR} \tilde{\bLambda}^{-\frac{1}{2}})^\top $, so we can use the $M\times N$ matrix $(\C \tilde{\RR} \tilde{\bLambda}^{-\frac{1}{2}})^\top$ as the new feature representation for view 1 (similarly for view 2), where inner products between samples approximate kernel similarities. We can extract such features for both views, and apply linear CCA to them to approximate the KCCA solution~\citep{Yang_12e,Lopez_14b}. Notice that using the Nystr\"om method has a time complexity (for view 1) of $\calO(M^2 d_x + M^3 + N M d_x + M^2 N)$, where the four terms account for the costs of forming $\tilde{\K}_x \in \bbR^{M\times M}$, computing the eigenvalue decomposition of $\tilde{\K}_x$, forming $\C$, and computing $\C \tilde{\RR} \tilde{\bLambda}^{-\frac{1}{2}}$, respectively, and a space complexity of $\calO(M^2)$ for saving the eigenvalue systems of $\tilde{\K}_x$ and $\tilde{\K}_y$,  which are expensive for large $M$. Although there have been various sampling/approximation strategies for the Nystr\"om method~\citep{Li_10b,ZhangKwok09a,ZhangKwok10a,Kumar_12a,GittenMahoney13a}, their constructions are more involved.

\subsection{Approximation via random features}

We now describe another approximate KCCA formulation that is particularly well-suited to large-scale problems.

It is known from harmonic analysis that a shift-invariant kernel of the form $k(\x,\x')=\kappa(\x-\x')$ is a positive definite kernel if and only if $\kappa(\Delta)$ is the Fourier transform of a non-negative measure (this is known as Bochner's theorem; see~\citealp{Rudin94a,RahimiRecht08a}). Thus we can write the kernel function as an expectation over sinusoidal functions over the underlying probability measures and approximate it with sample averages. Taking as a concrete example the Gaussian radial basis function (RBF) kernel $k(\x,\x')=e^{-\norm{\x-\x'}^2/2 s^2}$ where $s$ is the kernel width, it can be approximated as~\citep{Lopez_14b}
\begin{align*}
k(\x,\x') = \int e^{-j\w^\top(\x-\x')}p(\w) \ d \w \approx \frac{1}{M} \sum_{i=1}^M 2 \cos(\w_i^\top \x + b_i) \cos(\w_i^\top \x' + b_i),
\end{align*}
where $p(\w)$ is the multivariate Gaussian distribution $\calN(\0, \frac{1}{s^2} \I)$, obtained from the inverse Fourier transform of $\kappa(\Delta)=e^{-\frac{\norm{\Delta}^2}{2 s^2}}$,  and $b_i$ is drawn from a uniform distribution over $[0,2\pi]$. Approximations for other shift-invariant kernels (Laplacian, Cauchy) can be found in \citet{RahimiRecht08a}. This approach has been extended to other types of kernels~\citep{KarKarnic12a,Hamid_14a,Pennin_15a}. A careful quasi-Monte Carlo scheme for sampling from $p(\w)$~\citep{Yang_14a}, and structured feature transformation for accelerating the computation of $\w_i^\top \x$~\citep{Le_13a}, have also been studied.

Leveraging this result, \citet{RahimiRecht09a} propose to first extract $M$-dimensional \emph{random Fourier features} for input $\x$ as (with slight abuse of notation)
\begin{align*}
\phi(\x)=\sqrt{\frac{2}{M}} \left[\cos(\w_1^\top \x+b_1),\dots,\cos(\w_M^\top \x+b_M) \right] \in \bbR^M,
\end{align*}
so that $\phi(\x)^\top \phi(\x')\approx k(\x,\x')$, and then apply linear methods on these features.  
The computational advantage of this approach is that it turns nonlinear learning problems into convex linear learning problems, for which empirical risk minimization is much more efficient (e.g, \citealp{Lu_15b} used the recently proposed stochastic gradient method by \citealt{Leroux_12a} for the task of multinomial logistic regression with random Fourier features). \citet{RahimiRecht09a} showed that it allows us to effectively learn nonlinear models and still obtain good learning guarantees.

\citet{Lopez_14b} have recently applied the random feature idea to KCCA, by extracting $M$-dimensional random Fourier features $\{ (\phi_x(\x_i), \phi_y(\y_i)) \}_{i=1}^N$ for both views and solving exactly a linear CCA on the transformed pairs.  They also provide an approximation guarantee for this approach (see Theorem~4 of \citealp{Lopez_14b}).
Comparing random features with the Nystr\"om method described previously, when both techniques use rank-$M$ approximations, the cost of computing the solution to \eqref{e:cca} is the same and of order $\calO(M^2 N)$. But using random features, we generate the $M$-dimensional features in a data-independent fashion with a minimal cost $\calO(NMd_x)$ (for view 1), which is negligible compared to that of the Nystr\"om method. Furthermore, random features do not require saving any kernel matrix and the random features can be generated on the fly by saving the random seeds. Although 
the Nystr\"om approximation can be more accurate at the same rank~\citep{Yang_12e}, the computational efficiency and smaller memory cost of random features make them more appealing for large-scale problems in practice.

\section{Stochastic optimization of approximate KCCA}
\label{s:sto}

When the dimensionality $M$ of the random Fourier features is very large, solving the resulting linear CCA problem is still very costly as one needs to save the $M\times M$ matrix $\tilde{\T}=\tilde{\bSigma}_{xx}^{-\frac{1}{2}} \tilde{\bSigma}_{xy} \tilde{\bSigma}_{yy}^{-\frac{1}{2}}$ and compute its SVD, where the covariance matrices are now computed on $\{ (\phi_x(\x_i), \phi_y(\y_i)) \}_{i=1}^N$ instead of $\{ (\x_i, \y_i) \}_{i=1}^N$. It is thus desirable to develop memory-efficient stochastic optimization algorithms for CCA, where each update of the projection mappings depends only on a small minibatch of $b$ examples, thus reducing the memory cost to $\calO(bM)$. Notice, however, in contrast to the classification or regression objectives that are more commonly used with random Fourier features~\citep{RahimiRecht09a,Huang_14a,Lu_15b}, the CCA objective \eqref{e:cca} can not be written as an unconstrained sum or expectation of losses incurred at each training sample (in fact all training samples are coupled together through the constraints). As a result, stochastic gradient descent, which requires unbiased gradient estimates computed from small minibatches, is not directly applicable here.

Fortunately, \citet{Ma_15b,Wang_15c} have developed stochastic optimization algorithms, referred to as AppGrad (Augmented Approximate Gradient) and NOI (Nonlinear Orthogonal Iterations) respectively, for linear CCA and its deep neural network extension deep CCA~\citep{Andrew_13a}. Their algorithms are essentially equivalent other than the introduction in~\citep{Wang_15c} of a time constant for smoothing the covariance estimates over time.   The idea originates from the \emph{alternating least squares (ALS)} formulation of CCA~\citep{GolubZha95a,LuFoster14a}, which computes the SVD of $\T$ using orthogonal iterations (a generalization of power iterations to multiple eigenvalues/eigenvectors, \citealp{GolubLoan96a}) on $\T\T^\top$ and $\T^\top \T$. Due to the special form of $\T\T^\top$ and $\T^\top \T$, two least squares problems arise in this iterative approach (see, e.g., \citealp[Section~III. A]{Wang_15c} for more details). With this observation, \cite{LuFoster14a} solve these least squares problems using randomized PCA~\citep{Halko_11b} and a batch gradient algorithm. \citet{Ma_15b,Wang_15c} take a step further and replace the exact solutions to the least squares problems with efficient stochastic gradient descent updates. Although unbiased gradient estimates of these subproblems do not lead to unbiased gradient estimates of the original CCA objective, local convergence results (that the optimum of CCA is a fixed point of AppGrad, and the AppGrad iterate converges linearly to the optimal solution when started in its neighborhood) have been established for AppGrad~\citep{Ma_15b}. It has also been observed that the stochastic algorithms converge fast to approximate solutions that are on par with the exact solution or solutions by batch-based optimizers.

\begin{algorithm}[t]
  \caption{KNOI: Stochastic optimization for approximate KCCA.}
  \label{alg:sto}
  \renewcommand{\algorithmicrequire}{\textbf{Input:}}
  \renewcommand{\algorithmicensure}{\textbf{Output:}}
  \begin{algorithmic}
    \REQUIRE 
Initialization $\U \in \bbR^{M\times L}, \V \in \bbR^{M\times L}$, time constant $\rho$, minibatch size $b$, learning rate $\eta$, momentum $\mu$.
    \STATE $\Delta_{\U} \leftarrow \0$,\ \ \ \ $\Delta_{\V} \leftarrow \0$
    \STATE Randomly choose a minibatch $(\X_{b_0},\Y_{b_0})$ 
    \STATE $\SS_{xx} \leftarrow \frac{1}{\abs{b_0}} \sum_{i\in b_0} \left( \U^\top \phi_x(\x_i) \right) \left( \U^\top \phi_x(\x_i) \right)^\top $, 
    \STATE $\SS_{yy} \leftarrow \frac{1}{\abs{b_0}} \sum_{i\in b_0} \left( \V^\top \phi_y(\y_i) \right) \left( \V^\top \phi_y(\y_i) \right)^\top $
    \FOR{$t=1,2,\dots,T$}
    \STATE Randomly choose a minibatch $(\X_{b_t},\Y_{b_t})$ of size $b$
    \STATE $\SS_{xx} \leftarrow \rho \SS_{xx} + (1-\rho) \frac{1}{b} \sum_{i\in b_t} \left( \U^\top \phi_x(\x_i) \right) \left( \U^\top \phi_x(\x_i) \right)^\top$
    \STATE $\SS_{yy} \leftarrow \rho \SS_{yy} + (1-\rho) \frac{1}{b} \sum_{i\in b_t} \left( \V^\top \phi_y(\y_i) \right) \left( \V^\top \phi_y(\y_i) \right)^\top$
    \STATE Compute the gradient $\partial \U$ of the objective 
    \begin{gather*}
      \min_{\U}\; \frac{1}{b} \sum_{i\in b_t} \norm{ \U^\top \phi_x(\x_i) - \SS_{yy}^{-\frac{1}{2}} \V^\top \phi_y(\y_i) }^2
    \end{gather*}
    as $\partial \U \leftarrow \frac{1}{b} \sum_{i\in b_t} \phi_x(\x_i) \left(\U^\top \phi_x(\x_i) - \SS_{yy}^{-\frac{1}{2}} \V^\top \phi_y(\y_i) \right)^\top $
    \STATE Compute the gradient $\partial \V$ of the objective 
    \begin{gather*}
      \min_{\V}\; \frac{1}{b} \sum_{i\in b_t} \norm{ \V^\top \phi_y(\y_i) - \SS_{xx}^{-\frac{1}{2}} \U^\top \phi_x(\x_i) }^2
    \end{gather*}
    as $\partial \V \leftarrow \frac{1}{b} \sum_{i\in b_t} \phi_y(\y_i) \left(\V^\top \phi_y(\y_i) - \SS_{xx}^{-\frac{1}{2}} \U^\top \phi_x(\x_i) \right)^\top $
    \STATE $\Delta_{\U} \leftarrow \mu \Delta_{\U} - \eta \partial \U$,\ \ \ \ $\Delta_{\V} \leftarrow \mu \Delta_{\V} - \eta \partial \V$
    \STATE $\U \leftarrow \U + \Delta_{\U}$,\ \ \ \ $\V \leftarrow \V + \Delta{\V}$
    \ENDFOR  
    \ENSURE The updated $(\U, \V)$.
  \end{algorithmic}
\end{algorithm}

We give our stochastic optimization algorithm for approximate KCCA, named KNOI (Kernel Nonlinear Orthogonal Iterations), in Algorithm~\ref{alg:sto}. Our algorithm is adapted from the NOI algorithm of \citet{Wang_15c}, which allows the use of smaller minibatches (through the time constant $\rho$) than does the AppGrad algorithm of \citet{Ma_15b}. In each iteration, KNOI adaptively estimates the covariance of the projections of each view ($\in \bbR^L$) using a convex combination (with $\rho\in [0,1)$) of the previous estimate and the estimate based on the current minibatch,\footnote{In practice we also adaptively estimate the mean of the projections and center each minibatch.} uses them to whiten the targets of the cross-view least squares regression problems, derives gradients from these problems,\footnote{We also use small weight decay regularization ($\sim 10^{-5}$) for $(\U, \V)$ in the least squares problems.} and finally updates the projection matrices $(\U,\V)$ with momentum. Notice that $\rho$ controls how fast we forget the previous estimate; larger $\rho$ may be necessary for the algorithm to work well if the minibatch size $b$ is small (e.g., due to memory constraints), in which case the covariance estimates based on the current minibatch are noisier (see discussions in \citealp{Wang_15c}). Empirically, we find that using momentum $\mu\in[0, 1)$ helps the algorithm to make rapid progress in the objective with a few passes over the training set, as observed by the deep learning community~\citep{Sutskev_13a}. Although we have specifically use random Fourier features in Algorithm~\ref{alg:sto}, in principle other low-rank kernel approximations can be used as well. \karen{this comment should ideally be made much earlier, e.g. in the intro.  In general the point that the approach is more general than just for random Fourier features would be nice to make earlier.}

In each iteration of KNOI, the main cost comes from evaluating the random Fourier features and the projections for a minibatch, and computing the gradients. Since we usually look for low-dimensional projections ($L$ is small), it costs little memory and time to compute the covariance estimates $\SS_{xx}$ and $\SS_{yy}$ (of size $L\times L$) and their eigenvalue decompositions (for $\SS_{xx}^{-\frac{1}{2}}$ and $\SS_{yy}^{-\frac{1}{2}}$). Overall, KNOI has a memory complexity of $\calO(Mb)$ (excluding the $\calO(ML)$ cost for saving $\U$ and $\V$ in memory) and a time complexity of $\calO( b M (d_x+d_y+4L) )$ per iteration.

The $(\U,\V)$ we obtain from Algorithm~\ref{alg:sto} do not satisfy the constraints $\U^\top\tilde{\bSigma}_{xx}\U=\V^\top \tilde{\bSigma}_{yy}\V$; one can enforce the constraints via another linear CCA in $\bbR^L$ on $\{(\U^\top \phi_x(\x_i),\V^\top \phi_y(\y_i))\}_{i=1}^N$, which does not change the canonical correlation between the projections. To evaluate the projection of a view 1 test sample $\x$, we generate the random Fourier features $\phi_x(\x)$ using the same random seed for the training set, and compute $\U^\top \phi_x(\x)$ for it.

Finally, we comment on the choice of hyperparameters in KNOI. Empirically, we find that larger $b$ tends to give more rapid progress in the training objective, in which case $\rho$ can be set to small values or to $0$ as there is sufficient covariance information in a large minibatch (also shown by \citealp{Ma_15b,Wang_15c}). Therefore, we recommend using larger $b$ and $\rho=0$ if one can afford the memory cost. For large-scale problems with millions of training examples, we set $b$ to be a small portion of the training set (a few thousands) and enjoy the fast convergence of stochastic training algorithms~\citep{BottouBousquet08a}.  In our experiments we initialize $(\U,\V)$ with values sampled from a Gaussian distribution with standard deviation $0.1$, and tune the learning rate $\eta$ and momentum $\mu$ on small grids. 

\section{Related work}
\label{s:related}

There have been continuous efforts to scale up classical methods such as principal component analysis and partial least squares with stochastic/online updates~\citep{Krasul69a,OjaKarhun85a,WarmutKuzmin08a,Arora_12a,Arora_13a,Mitliag_13a,Balsub_13a,Shamir15a,Xie_15b}.  The CCA objective is more challenging due to the constraints, as also pointed out by~\citet{Arora_12a}.

\citet{Avron_13a} propose an algorithm for selecting a subset of training samples that retain the most information 
for accelerating linear CCA, when there are many more training samples (large $N$) than features (small $M$ in our case).  While this approach effectively reduces the training set size $N$, it provides no remedy for the large $M$ scenario we face in approximate KCCA.

In terms of online/stochastic CCA, \cite{Yger_12a} propose an adaptive CCA algorithm with efficient online updates based on matrix manifolds defined by the constraints (and they use a similar form of adaptive estimates for the covariance matrices).  However, the goal of their algorithm is anomaly detection for streaming data with a varying distribution, rather than to perform CCA for a given dataset. Regarding the stochastic CCA algorithms of \citet{Ma_15b,Wang_15c} we use here, an intuitively similar approach is proposed in the context of alternating conditional expectation~\citep{Makur_15a}.

Another related approach is that of \citet{Xie_15b}, who propose the Doubly Stochastic Gradient Descent (DSGD) algorithm for approximate kernel machines (including KCCA) based on random Fourier features.  KNOI and DSGD are different in several respects. First, the stochastic update rule of DSGD for $(\U, \V)$ is derived from the Lagrangian of an eigenvalue formulation of CCA and is different from ours, e.g., DSGD does not have any whitening steps while KNOI does. Second, DSGD gradually increases the number of random Fourier features (or cycles through blocks of random Fourier features) and updates the corresponding portions of $(\U, \V)$  as it sees more training samples. While this potentially further reduces the memory cost of the algorithm, it is not essential as we could also process the random Fourier features in minibatches (blocks) within KNOI.

\section{Experiments}
\label{s:expt}

In this section, we demonstrate the KNOI algorithm 
on two large-scale problems and compare it to several alternatives:
\begin{itemize}
\item CCA, solved exactly by SVD.
\item FKCCA, low-rank approximation of KCCA using random Fourier features, with the CCA step solved exactly by SVD.
\item NKCCA, low-rank approximation of KCCA using the Nystr\"om method, with the CCA step solved exactly by SVD.
\end{itemize}
We implement KNOI in MATLAB 
with GPU support. Since our algorithm mainly involves simple matrix operations, running it on a GPU provides significant speedup.

\subsection{MNIST 8M}

In the first set of experiments, we demonstrate the scability and efficiency of KNOI on the MNIST8M dataset~\citep{Loosli_07a}. The dataset consists of $8.1$ million $28\times 28$ grayscale images of the digits $0$-$9$. We divide each image into the left and right halves and use them as the two views in KCCA, so the input dimensionality is $392$ for both views. The dataset is randomly split into training/test sets of size 8M/0.1M. The task is to learn $L=50$ dimensional projections using KCCA, and the evaluation criterion is the total canonical correlation achieved on the test set (upper-bounded by $50$). The same task is used by \citet{Xie_15b}, although we use a different training/test split. 

As in \citet{Xie_15b}, we fix the kernel widths using the ``median'' trick\footnote{Following \citet{Xie_15b}, kernel widths are estimated from the median of pairwise distances between $4000$ randomly selected training samples.} 
for all algorithms. We vary the rank $M$ for FKCCA and NKCCA from $256$ to $6000$. For comparison, we use the same hyperparamters as those of \citet{Xie_15b}\footnote{We thank the authors for providing their MATLAB implementation of DSGD.}:  data minibatch size $b=1024$, feature minibatch size $2048$, total number of random Fourier features $M=20480$,\footnote{\citet{Xie_15b} used a version of random Fourier features with both $\cos$ and $\sin$ functions, so the number of learnable parameters (in $\U$ and $\V$) of DSGD is twice that of KNOI for the same $M$.} and a decaying step size schedule. For KNOI, we tune hyperparameters on a rough grid based on total canonical correlation obtained on a random subset of the training set with 0.1M samples, 
and set the minibatch size $b=2500$, time constant $\rho=0$, learning rate $\eta=0.01$, and momentum $\mu=0.995$. We run the iterative algorithms DSGD and KNOI for one pass over the data, so that they see the same number of samples as FKCCA/NKCCA. We run each algorithm $5$ times using different random seeds and report the mean results. 

\begin{table}[t]\centering
\caption{Total canonical correlation on MNIST 8M test set and the corresponding run times (with GPU run times in parentheses).}
\label{t:corr}
\begin{tabular}{l|r|c|r}\hline
Method & M & Canon. Corr. & Time (minutes)\\ \hline \hline
linear CCA &  & 26.8 & 0.5\hspace{2.7em} \\
\hline
      &  1024 & 33.5 & 9.8\hspace{2.7em} \\
      &  2048 & 37.6 & 24.4\hspace{2.7em} \\
FKCCA &  4096 & 40.7 & 68.9\hspace{2.7em} \\
      &  5000 & 41.4 & 79.4\hspace{2.7em} \\
      &  6000 & 42.1 & 107.5\hspace{2.7em} \\
\hline
      & 1024 & 39.6 & 17.1\hspace{2.7em} \\
NKCCA & 2048 & 42.2 & 44.9\hspace{2.7em} \\
      & 4096 & 44.1 & 138.8\hspace{2.7em} \\
      & 5000 & 44.5 & 196.3\hspace{2.7em} \\
      & 6000 & 44.8 & 272.2\hspace{2.7em} \\
\hline
DSGD & 20480 & 43.4 & 306.9\hspace{2.7em} \\
\hline
               & 20480   & 44.5  & 97.2\hspace{.75em}(7.6) \\
KNOI           & 40960   & 45.0  & 194.0 (15.4) \\
               & 100000  & 45.3  & 502.4 (39.7) \\
\hline
\end{tabular}
\end{table}

\begin{figure}
\centering
\psfrag{FKCCA M=4096}[l][l]{FKCCA $\ M=4096$}
\psfrag{NKCCA M=4096}[l][l]{NKCCA $M=4096$}
\psfrag{KNOI M=40960}[l][l]{KNOI $\ \ \ \ M=40960$}
\psfrag{samples}[][]{\# training samples ($\times 10^5$)}
 \psfrag{corr}[][]{Total Canon. Corr.}
\includegraphics[width=0.6\linewidth]{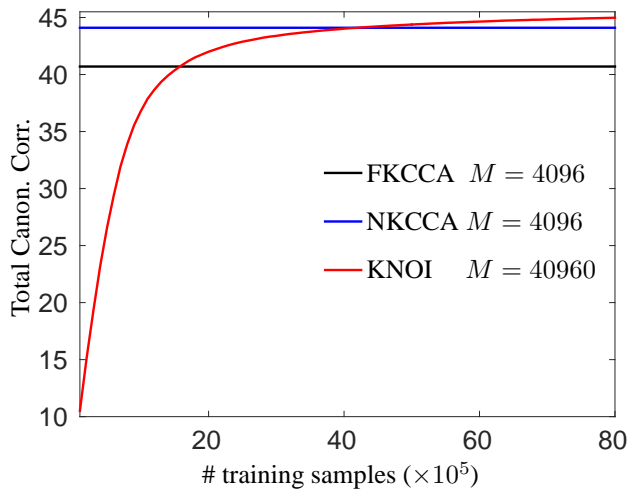}
\caption{Learning curve of KNOI on the MNIST 8M test set. We show total canonical correlation in the 50-dimensional projections vs.~the number of training samples ($\times 10^5$) processed by KNOI.  The FKCCA and NKCCA values, always obtained using the entire training set, are shown as horizontal lines.}
\label{f:learncurve}
\end{figure}

The total canonical correlations achieved by each algorithm on the test set, together with the run times measured on a workstation with $6$ $3.6GHz$ CPUs and $64G$ main memory, are reported in Table~\ref{t:corr}. As expected, all algorithms improve monotonically as $M$ is increased. FKCCA and NKCCA achieve competitive results with a reasonably large $M$, with NKCCA consistently outperforming FKCCA at the cost of longer run times. KNOI outperforms the other iterative algorithm DSGD, and overall achieves the highest canonical correlation with a larger $M$. We show the learning curve of KNOI with $M=40960$ (on the test set) in Figure~\ref{f:learncurve}. We can see that KNOI achieves steep improvement in the objective in the beginning, and already outperforms the exact solutions of FKCCA and NKCCA with $M=4096$ after seeing only 1/4 to 1/2 of the training set. 
We also run KNOI on an NVIDIA Tesla K40 GPU with 12G memory, and report the run times in parentheses in Table 1; the GPU provides a speedup of more than 12 times. For this large dataset, the KNOI algorithm itself requires less memory (less than 12G) than loading the training data in main memory ($\sim$25G).

\subsection{X-ray microbeam speech data}

In the second set of experiments, we apply approximate KCCA to the task of learning acoustic features for automatic speech recognition. We use the Wisconsin X-ray microbeam (XRMB) corpus~\citep{Westbur94a} of simultaneously recorded speech and articulatory measurements from 47 American English speakers. It has previously been shown that multi-view feature learning via CCA/KCCA greatly improves phonetic recognition performance given audio input alone~\citep{AroraLivesc13a,Wang_15a,Wang_15b}.

We follow the setup of \citet{Wang_15a,Wang_15b} and use the learned features (KCCA projections) for speaker-independent phonetic recognition.\footnote{Unlike \citet{Wang_15a,Wang_15b}, who used the HTK toolkit~\citep{Young_99a}, we use the Kaldi speech recognition toolkit~\citep{Povey_11a} for feature extraction and recognition with hidden Markov models.  Our results therefore don't match those in \citet{Wang_15a,Wang_15b} for the same types of features, but the relative merits of different types of features are consistent.}
The two input views are acoustic features (39D features consisting of mel frequency cepstral coefficients (MFCCs) and their first and second derivatives) and articulatory features (horizontal/vertical displacement of 8 pellets attached to several parts of the vocal tract) concatenated over a 7-frame window around each frame, giving 273D acoustic inputs and 112D articulatory inputs for each view.  The XRMB speakers are split into disjoint sets of 35/8/2/2 speakers for feature learning/recognizer training/tuning/testing. The 35 speakers for feature learning are fixed; the remaining 12 are used in a 6-fold experiment (recognizer training on 8 speakers, tuning on 2 speakers, and testing on the remaining 2 speakers). Each speaker has roughly $50K$ frames, giving 1.43M training frames for KCCA training. We remove the per-speaker mean and variance of the articulatory measurements for each training speaker. All of the learned feature types are used in a ``tandem'' speech recognition approach~\citep{Herman_00c}, i.e., they are appended to the original 39D features and used in a standard hidden Markov model (HMM)-based recognizer with Gaussian mixture observation distributions. 

For each fold, we select the hyperparameters based on recognition accuracy on the tuning set. For each algorithm, the feature dimensionality $L$ is tuned over $\{30, 50, 70\}$, and the kernel widths for each view are tuned by grid search. We initially set $M=5000$ for FKCCA/NKCCA, and also test FKCCA at $M=30000$ (the largest $M$ at which we could afford to obtain an exact SVD solution on a workstation with 64G main memory) with kernel widths tuned at $M=5000$; we could not obtain results for NKCCA with $M=30000$ in $48$ hours. For KNOI, we set $M=100000$ and tune the optimization parameters on a rough grid.  The tuned KNOI uses minibatch size $b=2500$, time constant $\rho=0$, fixed learning rate $\eta=0.01$, and momentum $\mu=0.995$. For this combination of $b$ and $M$, we are able to run the algorithm on a Tesla K40 GPU (with 12G memory), and each epoch (one pass over the 1.43M training samples) takes only $7.3$ minutes. We run KNOI for $5$ epochs and use the resulting acoustic view projection for recognition. We have also tried to run KNOI for $10$ epochs and the recognition performance does not change, even though the total canonical correlation keeps improving on both training and tuning sets. 

\begin{table}[t]
\centering
\caption{Mean phone error rates (PER) over 6 folds obtained by each algorithm on the XRMB test speakers.}
\label{t:xrmb_pers}
\begin{tabular}{l|c}
\hline
Method & Mean PER (\%)  \\
\hline \hline
Baseline (MFCCs) & 37.6 \\
CCA & 29.4 \\
FKCCA ($M=5000$) & 28.1 \\
FKCCA ($M=30000$)& 26.9 \\
NKCCA ($M=5000$) & 28.0 \\
KNOI ($M=100000$) & 26.4 \\
DCCA\ & \textbf{25.4} \\
\hline
\end{tabular}
\end{table}

For comparison, we report the performance of a baseline recognizer that uses only the original MFCC features, and the performance of deep CCA (DCCA) as described in \citet{Wang_15b}, which uses $3$ hidden layers of $1500$ ReLU units followed by a linear output layer in the acoustic view, and only a linear output layer in the articulatory view.  With this architecture, each epoch of DCCA takes about 8 minutes on a Tesla K40 GPU, on par with KNOI. Note that this DCCA architecture was tuned carefully for low PER rather than high canonical correlation. This architecture produces a total correlation of about $25$ (out of a maximum of $L=70$) on tuning data, while KNOI achieves $46.7$. DCCA using deeper nonlinear networks for the second view can achieve even better total canonical correlation, but its PER performance then becomes significantly worse.

Phone error rates (PERs) obtained by different algorithms 
are given in Table~\ref{t:xrmb_pers}, where smaller PER indicates better recognition performance. It is clear that all CCA-based features significantly improve over the baseline. Also, a large $M$ is necessary for KCCA to be competitive with deep neural network methods, which is consistent with the findings of \citet{Huang_14a,Lu_15b} when using random Fourier features for speech data (where the task is frame classification). Overall, KNOI outperforms the other approximate KCCA algorithms, although DCCA is still the best performer.

\section{Conclusion}

We have proposed kernel nonlinear orthogonal iterations (KNOI), a memory-efficient approximate KCCA algorithm based on random Fourier features and stochastic training of linear CCA. It scales better to large data and outperforms previous approximate KCCA algorithms in both the objective values (total canonical correlation) and running times (with GPU support).
   
It is straightforward to incorporate in our algorithm the faster random features of~\citet{Le_13a} which can be generated (for view 1) in time $\calO(N M \log d_x)$ instead of $\calO(N M d_x)$, or the Taylor features of \citet{Cotter_12a} which is preferable for sparse inputs, and random features for dot product or polynomial kernels~\citep{KarKarnic12a,Hamid_14a,Pennin_15a}, which have proven to be useful for different domains. It is also worth exploring parallelization and multiple kernel learning strategies of~\citet{Lu_15b} with random Fourier features to further bridge the gap between kernel methods and deep neural network methods.

Finally, as noted before, our algorithm does not use unbiased estimates of the gradient of the CCA objective. 
However, unbiased gradient estimates are not necessary for convergence of stochastic algorithms in general; a prominent example is the popular Oja's rule for stochastic PCA (see discussions in \citealp{Balsub_13a, Shamir15a}). Deriving global convergence properties for our algorithm is a challenging topic and the subject of ongoing work.

\subsubsection*{Acknowledgement}

This research was supported by NSF grant IIS-1321015.  The opinions expressed in this work are those of the authors and do not necessarily reflect the views of the funding agency.  The Tesla K40 GPUs used for this research were donated by NVIDIA Corporation.  We thank Bo Xie for providing his implementation of the doubly stochastic gradient algorithm for approximate KCCA, and Nati Srebro for helpful discussions.

\small
\bibliographystyle{iclr2016_conference}

\bibliography{iclr16b}
\end{document}